\documentclass[letterpaper,10pt,conference]{ieeeconf}
\IEEEoverridecommandlockouts

\usepackage{graphicx}
\usepackage{amsmath,amssymb}
\usepackage{booktabs}
\usepackage{multirow}
\usepackage{makecell}
\usepackage{tabularx}
\usepackage{array}
\usepackage{xcolor}
\usepackage[caption=false,font=footnotesize]{subfig}
\usepackage{cite}
\usepackage{algorithm}
\usepackage{algpseudocode}
\usepackage{hyperref}
\hypersetup{hidelinks}

\graphicspath{{figures/}}

\newcommand{\methodname}{\mbox{DiM-WAM}}

\title{\methodname{}: World Action Modeling with Diverse Historical Event Memory}
\author{%
\begin{tabular}{c}
Kai Wang$^{1,2}$, Zhaopeng Gu$^{1,2}$, Yixiang Chen$^{1}$, Yuan Xu$^{1}$, Qisen Ma$^{1}$,\\
Jiabing Yang$^{1}$, Zhaowen Li$^{2,*}$, Yan Huang$^{1,3,*}$, Liang Wang$^{1}$, Peng Su$^{2}$
\end{tabular}%
\thanks{$^{1}$Institute of Automation, Chinese Academy of Sciences (CASIA), Beijing, China. $^{2}$Shenzhen Yinwang Intelligent Technology Co., Ltd., Shenzhen, China. $^{3}$FiveAges, Beijing, China.}%
\thanks{$^{*}$Corresponding authors: Zhaowen Li and Yan Huang. E-mail: \texttt{lizhaowen@yinwang.com}; \texttt{yhuang@nlpr.ia.ac.cn}.}%
}

\begin{document}
\bstctlcite{BSTcontrol}
\maketitle

\begin{abstract}
World action models (WAMs) jointly predict future visual states and actions, but short local context limits temporally dependent tasks.
We introduce \methodname{}, which augments a WAM with Diverse Historical Event Memory (DHEM).
DHEM uses bank-conditioned candidate features, novelty-aware selection, and accumulated-mass-weighted fusion to retain complementary event tokens in bounded memory; these tokens condition video and action denoising, while auxiliary supervision encourages coarse progress cues.
In a training-matched comparison with LingBot-VA on RMBench, \methodname{} improves the average full-task success rate from \(34.8\%\) to \(69.8\%\).
Under the same demonstration and evaluation protocol on four real-world tasks, it improves the average stage success ratio from \(70.6\%\) to \(93.5\%\) and the average full-task success rate from \(52.5\%\) to \(90.0\%\).
Project page: \url{https://wangkai-casia.github.io/dim-wam/}.
\end{abstract}


\section{Introduction}

Vision-language-action models (VLAs) learn robot policies by predicting actions from language instructions and visual observations~\cite{arxiv2307.15818,arxiv2406.09246,arxiv2410.24164,arxiv2503.14734,arxiv2506.07961}. This action-centric formulation enables scalable policy learning, but sparse action labels provide limited supervision for fine-grained manipulation dynamics. WAMs alleviate this limitation by jointly predicting executable actions and future visual states, where future visual prediction provides dense temporal supervision~\cite{arxiv2410.06158,arxiv2601.21998,arxiv2602.15922}. However, existing WAMs remain limited in long-horizon tasks, where correct actions often cannot be inferred from current observations or short local contexts alone.

\begin{figure}[t]\centering
\includegraphics[width=0.98\linewidth]{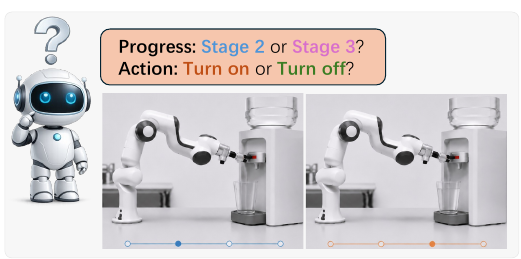}
\caption{Temporal ambiguity in the water-dispenser task: similar local observations near the switch require different actions depending on interaction history and task progress.}
\label{fig}
\end{figure}

Consider the water-dispenser task in Fig.~\ref{fig}. When the robot reaches for the same switch, the observations immediately before the turn-on and turn-off actions can appear visually similar, particularly when transparent water is difficult to perceive. Nevertheless, the required actions differ: the robot should interact with the switch at one stage but avoid or reverse the interaction at another. A WAM relying only on current observations or short contexts may therefore execute incorrect actions or stop at the wrong stage. Resolving this ambiguity requires access to key historical events and task-progress cues.

Long-horizon manipulation requires retaining different types of task-relevant events across tasks and stages. For example, some events indicate whether an interaction has occurred or a subtask has been completed, while others preserve information about layouts, target identities, or object-state changes needed later. Such diversity motivates a memory mechanism that can discover and preserve complementary event patterns.

Simply enlarging the local context increases computational cost and may dilute attention to sparse but task-critical events. Memory-augmented VLAs demonstrate the value of history~\cite{arxiv2501.18564,arxiv2508.19236,arxiv2603.03596,arxiv2603.12942}, but action-centric supervision provides limited signals for discovering and organizing heterogeneous historical events. In contrast, WAMs couple action learning with future visual prediction, providing richer supervision over both robot actions and environmental changes. This visual-action signal facilitates learning diverse historical event representations and enables the memory module to implicitly preserve complementary event patterns for prediction and control.

Inspired by the multiple memory subsystems (MMSS) view of human cognition, where functionally distinct memory systems may cooperate or compete~\cite{sherman2024multiple}, we organize historical information into multiple memory banks. Rather than assigning predefined semantics to individual banks, we allow them to specialize implicitly in complementary event patterns while sharing a bounded capacity. Based on this design, we propose \methodname{}, a memory-augmented WAM with Diverse Historical Event Memory (DHEM). DHEM preserves informative event tokens through novelty-aware selection, compresses redundancy via accumulated-mass-weighted fusion, and encourages task-stage representations through progress-aware training. The resulting memory tokens are combined with the local WAM context to condition video and action denoising.

Experiments on RMBench~\cite{chen2026rmbench} and four real-world Franka tasks show gains over the in-house WAM baselines, including a \(90.0\%\) average full-task success rate in the real-world evaluation.

\section{Related Work}

\subsection{WAMs}

Prior WAMs use future visual prediction to provide dense temporal supervision for robot policies. Early video-based approaches formulate decision making as conditional future-video generation and recover actions from predicted visual trajectories~\cite{arxiv2302.00111,arxiv2404.12377,arxiv2409.16283}. More recent methods strengthen visual-action coupling by jointly predicting future observations and actions, or unify action prediction with visual or proprioceptive state prediction and value estimation~\cite{arxiv2410.06158,arxiv2601.21998,arxiv2602.15922,arxiv2601.16163,arxiv2603.10448}. Efficiency-oriented variants compress predicted futures, separate joint video training from test-time generation, or decode only actions during inference~\cite{arxiv2602.22010,arxiv2603.16666,arxiv2603.17240}.

Existing WAMs improve future prediction, visual-action coupling, and inference efficiency, but do not address task-relevant history storage and retrieval under bounded memory. Our work complements them by introducing observation-grounded long-term memory into WAMs.

\subsection{Memory-Augmented Vision-Language-Action Models}

Memory-augmented VLAs incorporate historical information into policy inference through several complementary designs. SAM2Act reuses object-centric visual traces~\cite{arxiv2501.18564}, while MemoryVLA separates perceptual and cognitive memory for low-level evidence and high-level task context~\cite{arxiv2508.19236}. Other methods organize history across multiple temporal scales~\cite{arxiv2603.03596}, maintain context through recurrent queries~\cite{arxiv2603.12942}, selectively retain event-driven evidence~\cite{arxiv2606.20092}, or combine memory with imagination-based temporal modeling~\cite{arxiv2606.09827}.

These methods improve temporal reasoning in VLA policies, but primarily target action inference rather than memory design for joint prediction and control in WAMs.

\subsection{Memory-Augmented Long Video Generation}

Long-video generation faces a related challenge: extending fixed-context models to longer horizons while preserving temporal consistency. Existing methods improve sequence efficiency through state-space models, block scans, or linear attention~\cite{arxiv2403.07711,arxiv2505.20171,arxiv2509.24695}. Others compress or organize history with recurrent memory vectors, key-value (KV) cache compression, local-global caches, retrieval, or entity-centric memory~\cite{arxiv2502.12632,arxiv2512.04519,arxiv2603.21366,arxiv2512.14699,arxiv2601.03655,arxiv2605.31033}.

Long-video memory mainly targets visual consistency, whereas WAM memory must encode task progress, object states, and action consequences for closed-loop control.


\section{Method}

\subsection{Overview: Multi-Scale World Action Modeling}
\label{sec:method_overview}
\label{sec:multi_scale_wam}

\paragraph{DiM-WAM framework}

\begin{figure*}[t]
  \centering
  \includegraphics[width=0.95\textwidth]{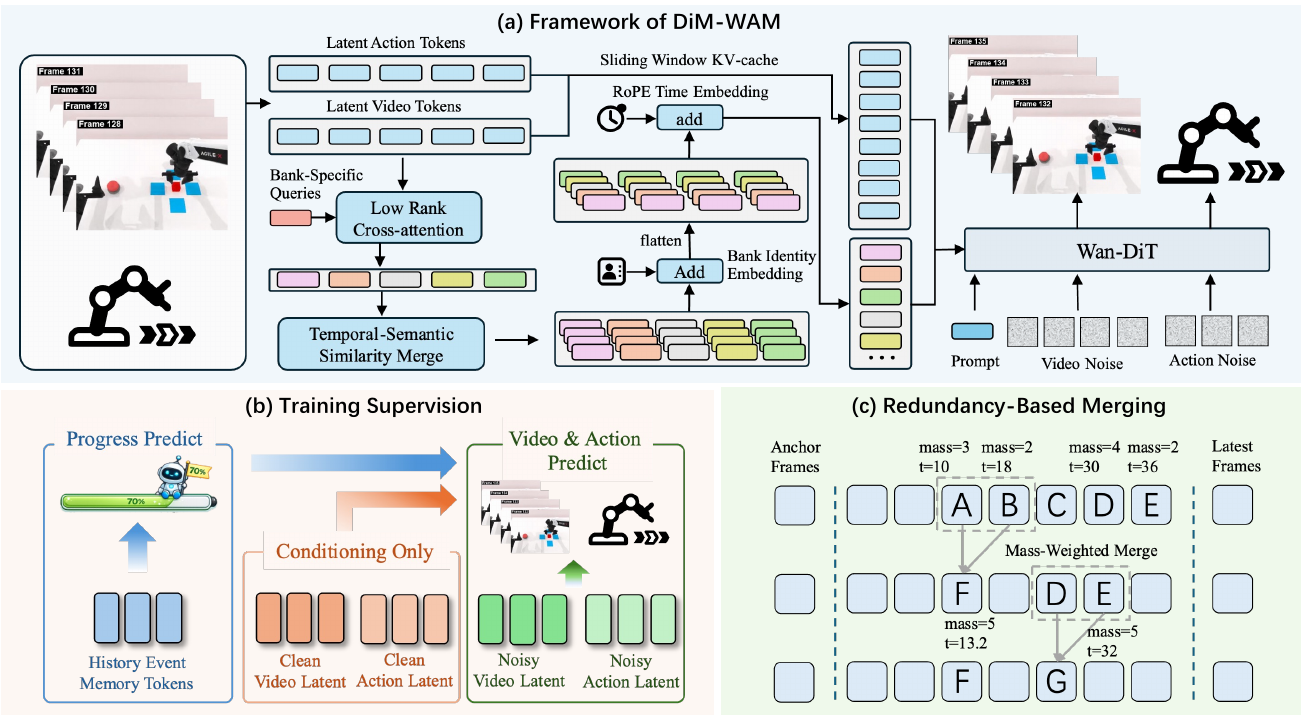}
  \caption{Overview of \methodname{}: (a) the DHEM-conditioned WAM framework, (b) task-progress supervision, and (c) redundancy-based memory merging.}
  \label{fig:framework_supervision_merge}
\end{figure*}

As shown in Fig.~\ref{fig:framework_supervision_merge}(a), at decision segment \(i\), \methodname{} takes language instruction \(c\), local key--value (KV) context \(\mathcal C_i\) from the sliding-window WAM cache, and persistent long-term memory state \(\mathcal M_i\). The valid event tokens in \(\mathcal M_i\) form the readable memory-token sequence \(\mathbf R_i\) defined in Sec.~\ref{sec:cooperative_memory_reading}. The model predicts future visual latents and actions over prediction horizon \(H\). The memory-token sequence conditions both denoising branches, while an auxiliary progress objective supervises the memory readout during training.

For bookkeeping, we summarize the temporal evidence available to the model as
\begin{equation}
\mathcal I_i = \left(\mathcal H_i^{\mathrm{short}},\mathcal H_i^{\mathrm{long}},\mathcal F_i^{\mathrm{short}},\mathcal G_i^{\mathrm{prog}}\right),
\label{eq:temporal_evidence_decomposition}
\end{equation}
where \(\mathcal I_i\) collects short-term history from the sliding-window KV cache, long-term history from DHEM, predicted short-term video-action evolution, and coarse task-progress information.

During inference, \methodname{} predicts future visual latents and actions over horizon \(H\) from the language instruction, local context, and long-term memory:
\begin{equation}
\left(\hat{\mathbf z}_{i:i+H-1},\hat{\mathbf a}_{i:i+H-1}\right)
= f_\theta\!\left(\mathcal C_i, c;\mathcal M_i\right),
\label{eq:wam_conditioned_prediction}
\end{equation}
where \(f_\theta\) maps the local context, instruction, and the readable representation of \(\mathcal M_i\) to future visual latents \(\hat{\mathbf z}\) and actions \(\hat{\mathbf a}\). The local context captures recent motion and visual cues, while the persistent memory state provides cross-stage evidence such as past interactions and object-state changes.

\subsection{Diverse Historical Event Memory Banks}
\label{sec:dhem}

Historical events in long-horizon tasks have different semantic roles and retention needs. If all events share one bank, heterogeneous but temporally adjacent events compete for slots and may be destructively mixed by similarity-based merging. We therefore use multiple parallel memory banks to maintain complementary evidence.

Before decision segment \(i\), the long-term memory state of DHEM is denoted by \(\mathcal M_i\).
It consists of \(K\) parallel memory banks, and each bank contains \(N\) bounded slots:
\begin{equation}
\begin{gathered}
\mathcal M_i = \left\{\mathcal M_i^k\right\}_{k=1}^{K}, \qquad
\mathcal M_i^k = \left\{e_{i,k,n}\right\}_{n=1}^{N},\\
e_{i,k,n} = \left(\mathbf m_{i,k,n},t_{i,k,n},\alpha_{i,k,n},v_{i,k,n}\right),
\end{gathered}
\label{eq:memory_state}
\end{equation}
where \(K\) is the number of memory banks, \(N\) is the number of slots per bank, \(k\) indexes banks, and \(n\) indexes slots. The vector \(\mathbf m_{i,k,n}\in\mathbb R^d\) is an event token with dimension \(d\), \(t_{i,k,n}\) is its timestamp, \(\alpha_{i,k,n}\) is its accumulated mass, and \(v_{i,k,n}\in\{0,1\}\) is its validity indicator. The tuple \(e_{i,k,n}\) is a memory event, and \(\mathcal M_i^k\) is one bounded memory bank. Thus, each memory event stores a token, timestamp, accumulated mass, and validity, and total memory cost is \(O(KN)\).

Bank roles are not predefined. Learnable bank-specific summary queries induce different observation views during writing, bank-local maintenance yields different retention trajectories, and the diversity loss discourages collapse. During reading, bank identity embeddings mark token sources.

\subsection{Writing and Maintenance of Historical Events}
\label{sec:memory_writing_maintenance}

After extracting a bank-conditioned candidate event feature from the current observation collected from the environment, each bank independently updates its bounded memory using novelty-aware retention and redundancy-based compression.

\paragraph{Visual event feature extraction and writing}
After segment \(i\), the current observation collected from the environment is encoded as visual features \(\mathbf X_i=[\mathbf x_{i,j}]_{j=1}^{L_i}\), where \(L_i\) is the number of visual tokens in segment \(i\). Each bank uses its learnable summary query \(\mathbf q_k\) to extract a candidate event feature. The normalized observation features and bank query are
\begin{equation}
\mathbf h_{i,j}=\operatorname{LN}_{x}(\mathbf x_{i,j}),\qquad
\mathbf r_k=\operatorname{LN}_{q}(\mathbf q_k),
\label{eq:bank_normalization}
\end{equation}
Using shared query and key projections with projection dimension \(d_r\), bank \(k\) extracts
\begin{equation}
\begin{aligned}
a_{i,k,j}
&=\operatorname{Softmax}_{j}\!\left(
\frac{(W_q\mathbf r_k)^\top(W_h\mathbf h_{i,j})}{\sqrt{d_r}}
\right)\\
\mathbf s_{i,k}
&= \sum_{j=1}^{L_i}a_{i,k,j}\mathbf h_{i,j}\\
\tilde{\mathbf u}_{i,k}
&=\mathbf s_{i,k}+\operatorname{FFN}\!\left(\operatorname{LN}_{o}(\mathbf s_{i,k})\right),
\end{aligned}
\label{eq:bank_conditioned_extraction}
\end{equation}
where \(a_{i,k,j}\) are token weights, \(\mathbf s_{i,k}\) is the attended summary, and \(\tilde{\mathbf u}_{i,k}\) is the candidate event feature. LN denotes layer normalization and FFN denotes a feed-forward network. All banks observe the same observation features but produce different candidate event features; persistent memory stores event tokens derived from these features, not generated predictions.

\paragraph{Novelty-aware bounded event compression}
In each bank, slot \(1\) is a fixed initial-state anchor, slot \(N\) is the latest event, and slots \(2,\ldots,N-1\) contain compressed history. Omitting segment and bank indices, the redundancy priority for two valid events \(p,q\) is
\begin{equation}
\rho_k(p,q)=\frac{1+\cos(\mathbf m_p,\mathbf m_q)}{2}\exp\!\left(-\frac{|t_p-t_q|}{\tau}\right),
\label{eq:merge_priority}
\end{equation}
where \(\tau>0\) is the temporal-decay constant. High priority indicates nearby semantic redundancy, i.e., low relative novelty.

When a memory bank is full, only adjacent middle-history pairs are eligible:
\begin{equation}
\mathcal A_{i,k}=\left\{(s,s+1)\mid 2\le s\le N-2,\ v_{k,s}v_{k,s+1}=1\right\}.
\label{eq:eligible_adjacent_pairs}
\end{equation}
The most redundant merge candidate among existing middle-history pairs is
\begin{equation}
(p^\star,q^\star)=\arg\max_{(p,q)\in\mathcal A_{i,k}}\rho_k(p,q),
\label{eq:selected_merge_pair}
\end{equation}
and the redundancy between the incoming candidate and the latest event is
\begin{equation}
\rho^{\mathrm{new}}_{i,k}=
\frac{1+\cos(\mathbf m_{k,N},\tilde{\mathbf u}_{i,k})}{2}
\exp\!\left(-\frac{|t_{k,N}-i|}{\tau}\right),
\label{eq:new_event_redundancy}
\end{equation}
where \(\mathcal A_{i,k}\) is the eligible adjacent-pair set, \((p^\star,q^\star)\) is the selected pair, and \(\rho^{\mathrm{new}}_{i,k}\) measures incoming-candidate redundancy.
The discard condition is
\begin{equation}
\rho^{\mathrm{new}}_{i,k}\ge \rho_k(p^\star,q^\star).
\label{eq:adaptive_write_threshold}
\end{equation}
The incoming candidate is therefore discarded when its redundancy with the latest event is no smaller than the maximum redundancy among eligible historical pairs. In this case, retaining the candidate would require sacrificing a less redundant historical pair. Otherwise, \((p^\star,q^\star)\) is merged to make room for the candidate.

The merged event preserves accumulated evidence through mass-weighted fusion:
\begin{equation}
\begin{gathered}
\mathbf m_{pq}
=\frac{\alpha_p\mathbf m_p+\alpha_q\mathbf m_q}
{\alpha_p+\alpha_q},\qquad
t_{pq}
=\frac{\alpha_pt_p+\alpha_qt_q}
{\alpha_p+\alpha_q},\\
\alpha_{pq}=\alpha_p+\alpha_q,
\end{gathered}
\label{eq:mass_merge}
\end{equation}
where \(\mathbf m_{pq}\), \(t_{pq}\), and \(\alpha_{pq}\) are the token, timestamp, and accumulated mass of the merged memory event. The mass-weighted timestamp \(t_{pq}\) may be noninteger and is interpreted as a continuous representative time in Eq.~\eqref{eq:merge_priority}. Because it lies between the timestamps of the adjacent inputs, compacting after the merge preserves chronological order. This update is illustrated in Fig.~\ref{fig:framework_supervision_merge}(c).

Algorithm~\ref{alg:memory_update} summarizes one bank update. We use the same similarity function and \(\tau=N-1\) in every bank; banks diverge because Eq.~\eqref{eq:bank_conditioned_extraction} supplies bank-specific candidates and updates depend on each bank's retained state. Tensor initialization fills all slots with valid copies of the first memory event. A slot is an initialization duplicate of the fixed anchor precisely when its timestamp equals the anchor timestamp, \(t_r=t_1\); these copies are evicted before the normal similarity-based rule.

\begin{algorithm}[t]
\caption{Update of bank \(k\) at segment \(i\)}
\label{alg:memory_update}
\begin{algorithmic}[1]
\Require Bank \(\mathcal M^k\), features \(\mathbf X_i\), query \(\mathbf q_k\), time \(i\)
\Ensure Updated bank \(\mathcal M^k\)
\State \(\mathbf r_k\gets\operatorname{LN}_{q}(\mathbf q_k)\)
\State \(\tilde{\mathbf u}_{i,k}\gets\operatorname{BankSummarize}(\mathbf X_i;\mathbf r_k)\) using Eq.~\eqref{eq:bank_conditioned_extraction}
\State \(e_{\mathrm{new}}\gets(\tilde{\mathbf u}_{i,k},i,1,1)\)
\If{\(\mathcal M^k\) is uninitialized}
  \State Fill all slots with \(e_{\mathrm{new}}\); fix \(e_1\) as the anchor
  \State \Return \(\mathcal M^k\)
\EndIf
\If{\(t_N=t_1\) (the latest slot is an initialization duplicate)}
  \State \(e_N\gets e_{\mathrm{new}}\)
  \State \Return \(\mathcal M^k\)
\EndIf
\If{a middle slot \(e_r\) satisfies \(t_r=t_1\)}
  \State Remove \(e_r\), compact the middle slots, and move \(e_N\) to slot \(N-1\)
  \State \(e_N\gets e_{\mathrm{new}}\)
  \State \Return \(\mathcal M^k\)
\EndIf
\If{an invalid middle slot exists}
  \State Move \(e_N\) to the first invalid middle slot
  \State \(e_N\gets e_{\mathrm{new}}\)
  \State \Return \(\mathcal M^k\)
\EndIf
\State Compute the highest-priority adjacent pair in the middle history:
\State \((p^\star,q^\star)\gets
\arg\max_{(p,q)\in\mathcal A_{i,k}}\rho_k(p,q)\)
\If{\(\rho^{\mathrm{new}}_{i,k}\ge\rho_k(p^\star,q^\star)\)}
  \State Keep the previous latest event \(e_N\) unchanged
  \State \Return \(\mathcal M^k\)
\EndIf
\State Save \(e_N\) and merge \(e_{p^\star},e_{q^\star}\) using Eq.~\eqref{eq:mass_merge}
\State Compact the middle slots in chronological order
\State Move the saved latest event to slot \(N-1\)
\State \(e_N\gets e_{\mathrm{new}}\)
\State \Return \(\mathcal M^k\)
\end{algorithmic}
\end{algorithm}

Each bank applies this rule to its own retained history and candidate event feature. Slot~\(1\) remains the anchor, slot~\(N\) remains the latest retained event, and slots~\(2,\ldots,N-1\) remain chronologically ordered middle history after every return path. Memory cost is \(O(KN)\), independent of trajectory length.

\subsection{Cooperative Memory Reading}
\label{sec:cooperative_memory_reading}

After bank-local writing, valid event tokens are flattened in slot-major, bank-minor order. Each token is augmented with a bank-identity embedding \(\mathbf b_k\), which encodes its source bank. A rotary position embedding (RoPE)~\cite{su2021roformer} injects slot-relative temporal order without changing persistent stored content. For coordinate assignment, we reindex each bank's slots by zero-indexed \(n\in\{0,\ldots,N-1\}\):
\begin{equation}
\pi_{i,n}=\pi_i^{\mathrm{cur}}-2(N-1-n),
\label{eq:slot_temporal_coordinate}
\end{equation}
where \(\pi_i^{\mathrm{cur}}\) is the RoPE coordinate of the current decision segment. The step of two coordinate units is a fixed implementation convention rather than an additional learned temporal scale. All banks at the same slot share \(\pi_{i,n}\); the bank identity embedding \(\mathbf b_k\) distinguishes their sources. Slot-major, bank-minor flattening therefore groups tokens by slot coordinate and then orders banks within each group. The readout is
\begin{equation}
\mathbf R_i=
\operatorname{RoPE}_{\pi}
\left(
\operatorname{Flat}_{n,k}
\left[
\left(\mathbf m_{i,k,n}+\mathbf b_k,\pi_{i,n}\right)
\right]_{v_{i,k,n}=1}
\right),
\label{eq:memory_readout}
\end{equation}
where \(\mathbf R_i\) is the Transformer-readable memory-token sequence. Slot recency, bank source, and event content are jointly visible during prediction, while writing remains bank-local. RoPE follows retained slot order rather than reordering memory events by their mass-weighted timestamps.

\begin{figure}[t]
  \centering
  \includegraphics[width=0.76\linewidth]{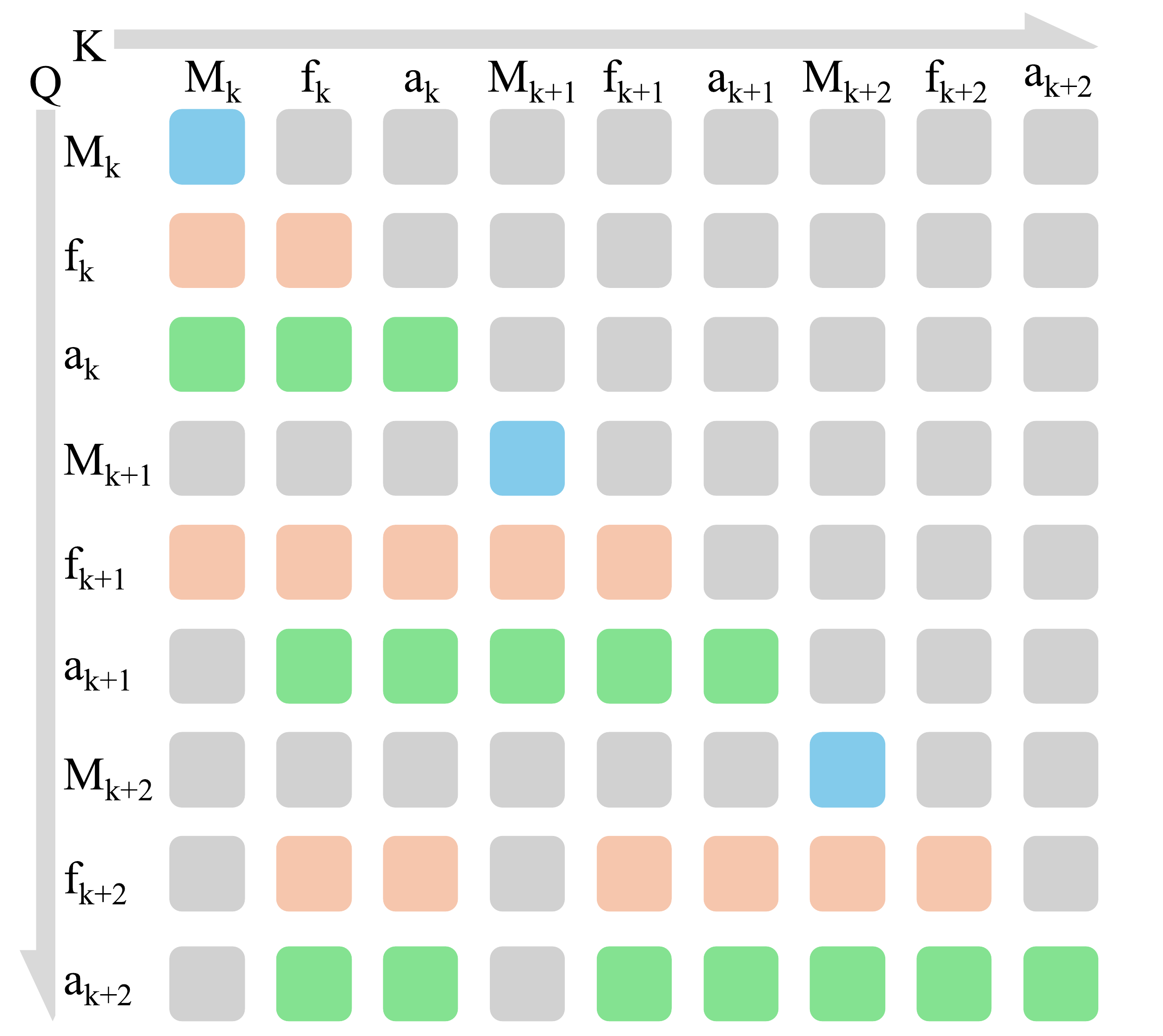}
  \caption{Visibility among memory, video-latent, and action tokens during denoising.}
  \label{fig:mask}
\end{figure}

As illustrated in Fig.~\ref{fig:mask}, event tokens condition current denoising but are not appended to the persistent KV cache; persistent memory is updated only from subsequent observations collected from the environment.

\subsection{Task-Progress-Aware Training}
\label{sec:progress_training}

As illustrated in Fig.~\ref{fig:framework_supervision_merge}(b), an auxiliary head predicts a coarse progress bin during training. This supervision encourages memory representations to encode trajectory-progress cues without treating the head as an explicit planner:
\begin{equation}
\mathbf p_i=\operatorname{Softmax}\!\left(h_{\mathrm{prog}}\left(\operatorname{Pool}(\mathbf R_i)\right)\right),\qquad
\mathbf p_i\in\Delta^{B-1},
\label{eq:progress_prediction}
\end{equation}
where \(h_{\mathrm{prog}}\) is the progress head, \(B\) is the number of progress bins, \(\Delta^{B-1}\) is the probability simplex, and \(\operatorname{Pool}\) denotes mean pooling over valid event tokens. For a demonstration containing \(T\) decision segments with zero-indexed \(i\in\{0,\ldots,T-1\}\), the normalized completion ratio is safely defined as
\begin{equation}
r_i=
\begin{cases}
\dfrac{i}{T-1}, & T>1,\\
0, & T=1
\end{cases}.
\label{eq:progress_ratio}
\end{equation}
The target progress bin \(y_i\) and the loss are
\begin{equation}
\begin{aligned}
y_i
&=\min\!\left(\left\lfloor Br_i\right\rfloor,B-1\right),\\
\mathcal L_{\mathrm{prog}}
&=-\log p_{i,y_i},
\end{aligned}
\label{eq:progress_loss}
\end{equation}
where \(p_{i,y_i}\) is the predicted probability of the target bin. The progress head supplies only an auxiliary training objective and does not participate in the inference-time decision path. It encourages retained memory events to encode coarse trajectory-progress cues rather than planning long-horizon behavior.

\paragraph{Joint objective}

To discourage bank collapse, let \(\mathcal V_{i,k}=\{n\mid v_{i,k,n}=1\}\) and define the normalized bank mean
\begin{equation}
\boldsymbol\mu_{i,k}=\frac{1}{|\mathcal V_{i,k}|}\sum_{n\in\mathcal V_{i,k}}\mathbf m_{i,k,n},\qquad
\bar{\boldsymbol\mu}_{i,k}=\frac{\boldsymbol\mu_{i,k}}{\|\boldsymbol\mu_{i,k}\|_2},
\label{eq:bank_mean}
\end{equation}
where \(\mathcal V_{i,k}\) is the valid-slot set and \(\bar{\boldsymbol\mu}_{i,k}\) is the normalized bank mean. 
For \(\mathcal P=\{(k,l)\mid 1\le k<l\le K\}\), the diversity loss is
\begin{equation}
\mathcal L_{\mathrm{div}}=\frac{1}{|\mathcal P|}\sum_{(k,l)\in\mathcal P}
\left(\bar{\boldsymbol\mu}_{i,k}^{\top}\bar{\boldsymbol\mu}_{i,l}\right)^2.
\label{eq:diversity_loss}
\end{equation}
Minimizing squared cosine similarity discourages bank collapse while allowing roles to emerge from bank-specific writing queries and state-dependent maintenance.

The complete objective is
\begin{equation}
\mathcal L=\mathcal L_{\mathrm{video}}+\mathcal L_{\mathrm{action}}
+\lambda_{\mathrm{div}}\mathcal L_{\mathrm{div}}+\lambda_{\mathrm{prog}}\mathcal L_{\mathrm{prog}}.
\label{eq:complete_objective}
\end{equation}
\(\mathcal L_{\mathrm{video}}\) is the diffusion denoising mean-squared error on future visual latents, and \(\mathcal L_{\mathrm{action}}\) is the corresponding masked diffusion denoising mean-squared error for action prediction.
The video and action losses in Eq.~\eqref{eq:complete_objective} have unit weights, while \(\lambda_{\mathrm{div}}\) and \(\lambda_{\mathrm{prog}}\) balance the two auxiliary objectives.


\section{Experiments}

We evaluate temporally dependent long-horizon tasks in simulation and on real-world robots, analyze the roles of DHEM and progress supervision, and investigate whether the banks learn distinct behaviors.

\subsection{Experimental Setup}

\paragraph{Base model and benchmark}
\methodname{} extends LingBot-VA~\cite{arxiv2601.21998} while preserving its original video-and-action input--output interface, and augments it with DHEM.

\begin{figure*}[t]
  \centering
  \includegraphics[width=0.8\textwidth]{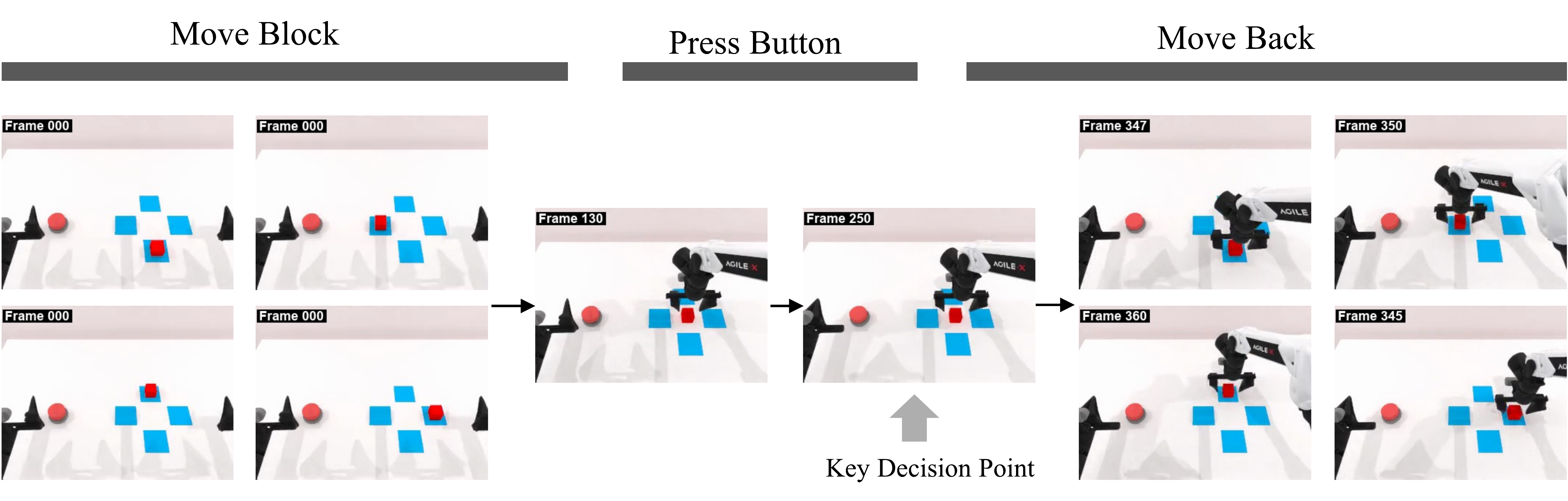}
  \caption{Key frames of the RMBench \texttt{put\_back\_block} task. Similar intermediate observations may require different final target locations depending on the initial state.}
  \label{fig:rmbench_task}
\end{figure*}

RMBench~\cite{chen2026rmbench}, built on RoboTwin~2.0, includes five \(M(1)\) tasks that rely on a few key historical observations and four \(M(n)\) tasks that require history accumulated across interaction steps. As shown in Fig.~\ref{fig:rmbench_task}, in \texttt{put\_back\_block}, the robot moves a block to the center, presses a button, and then returns it to its original location, which is one of four possible locations. Different initial states can yield similar intermediate observations, so the policy may fail if it does not retain the relevant historical event.

\subsection{Evaluation-Protocol Audit}

Table~\ref{tab:protocol_audit} audits whether camera views and temporal coverage expose the initial location to the local policy on \texttt{put\_back\_block}. Here, the window is the number of input frames, and stride is the input-frame subsampling interval rather than the action horizon. The audit uses LingBot-VA with its official training-parameter configuration and varies only the camera views, window, and stride.

\begin{table}[t]
\centering
\caption{Protocol audit for the RMBench \texttt{put\_back\_block} task. ``Chance'' denotes the ideal target-choice accuracy based only on locally visible target information and is not a full-task upper bound.}
\label{tab:protocol_audit}
\small
\renewcommand{\arraystretch}{1.08}
\setlength{\tabcolsep}{2pt}
\resizebox{\linewidth}{!}{
\begin{tabular}{@{}lccccc@{}}
\toprule
Camera views & Window & Stride & \makecell{Initial state\\visible} & \makecell{Chance\\(\%)} & \makecell{Success\\(\%)} \\
\midrule
front + wrist + head & 30  & 1 & No  & 25  & 86 \\
head + front         & 30  & 1 & No  & 25  & 41 \\
head + front         & 60  & 1 & No  & 25  & 49 \\
head + front         & 120 & 1 & Yes & 100 & 88 \\
front + wrist + head & 30  & 4 & Yes & 100 & \textbf{100} \\
head + front         & 30  & 4 & Yes & 100 & 98 \\
\bottomrule
\end{tabular}}
\end{table}

The wrist view raises success from \(41\%\) to \(86\%\) despite the initial state lying outside the local window, suggesting pose-dependent leakage. We therefore use head + front views, a 30-frame window, and stride 1 for the \(M(1)\) runs marked \(\dagger\) in Table~\ref{tab:rmbench_main}. As the audit covers only \texttt{put\_back\_block}, residual leakage in other tasks cannot be ruled out.

\paragraph{Implementation details}
Following the protocol audit, Fast-WAM, LingBot-VA, and \methodname{} use the same audit-guided RMBench protocol: head + front views, a 30-frame window, \(128\times128\) inputs, and strides 1 and 2 for \(M(1)\) and \(M(n)\), respectively. The \(\dagger\) columns in Table~\ref{tab:rmbench_main} denote these protocol-matched evaluations.

Within this protocol, LingBot-VA and \methodname{} form the training-matched controlled comparison, sharing 50 demonstrations, non-method optimization settings, 1,500 training steps, and 100 evaluation rollouts per task. Only the proposed DHEM and progress-supervision components and their objective terms differ. Fast-WAM~\cite{arxiv2603.16666} follows its official optimization recipe and serves as an additional protocol-matched in-house WAM baseline.

\methodname{} uses AdamW with learning rate \(10^{-5}\), batch size 1, gradient accumulation over 10 steps, and 1,500 optimization steps. DHEM uses $\lambda_{\mathrm{div}}=10^{-3}$, $\lambda_{\mathrm{prog}}=10^{-2}$, eight 12-slot banks, $B=10$, clean-KV dropout of $0.3$, and four-segment truncated backpropagation. Each task-specific model is trained on eight NVIDIA H800 GPUs; comparisons are not compute matched.

Real-robot experiments use one third-person camera at \(224\times224\); LingBot-VA and \methodname{} otherwise retain the simulation configuration. Fast-WAM and the \(\pi_{0.5}\) baseline~\cite{arxiv2504.16054} use their recommended recipes of 150 epochs and 20,000 optimization steps, respectively. All methods predict seven absolute joint positions and one binary gripper state.

\subsection{Simulation Experiments}

Among the in-house WAMs, Fast-WAM, LingBot-VA, and \methodname{} are evaluated under the audit-guided protocol. LingBot-VA and \methodname{} provide the training-matched controlled comparison for isolating the effect of DHEM and progress supervision, while Fast-WAM serves as an additional protocol-matched WAM baseline. Separately, DP~\cite{arxiv2303.04137}, ACT~\cite{arxiv2304.13705}, \(\pi_{0.5}\)~\cite{arxiv2504.16054}, X-VLA~\cite{arxiv2510.10274}, and Mem-0~\cite{chen2026rmbench} are RMBench-reported results included only as contextual reference points. Only the LingBot-VA versus \methodname{} comparison supports attribution to the proposed memory design.

\begin{table*}[t]
\centering
\caption{Main simulation results on RMBench long-horizon tasks. Values are success rates in percent. TMC denotes the RMBench \(M(1)\)/\(M(n)\) task-memory category. DP, ACT, \(\pi_{0.5}\), X-VLA, and Mem-0 are benchmark-reported results. The \(\dagger\) columns are protocol-matched in-house WAM evaluations; LingBot-VA and \methodname{} additionally form a training-matched comparison.}
\label{tab:rmbench_main}
\resizebox{0.85\textwidth}{!}{
\begin{tabular}{llccccccccc}
\toprule
Task & TMC & DP & ACT & \(\pi_{0.5}\) & X-VLA & Mem-0 & Fast-WAM\(\dagger\) & LingBot-VA\(\dagger\) & \methodname{}\(\dagger\) \\
\midrule
Observe and Pick Up & \(M(1)\) & 1.0 & 1.0 & 9.0 & 9.0 & 4.0 & 2.0 & 4.0 & \textbf{13.0} \\
Rearrange Blocks & \(M(1)\) & 0.0 & 29.0 & 13.0 & 13.0 & 89.0 & 0.0 & 63.0 & \textbf{99.0} \\
Put Back Block & \(M(1)\) & 0.0 & 0.0 & 11.0 & 18.0 & 90.0 & 0.0 & 41.0 & \textbf{98.0} \\
Swap Blocks & \(M(1)\) & 11.0 & 2.0 & 24.0 & 16.0 & 67.0 & 3.0 & 38.0 & \textbf{96.0} \\
Swap T & \(M(1)\) & 20.0 & 2.0 & 15.0 & 3.0 & 14.0 & 9.0 & 25.0 & \textbf{97.0} \\
\(M(1)\) Average & -- & 6.4 & 6.8 & 14.4 & 11.8 & 52.8 & 2.8 & 34.2 & \textbf{80.6} \\
\midrule
Battery Try & \(M(n)\) & 10.0 & 19.0 & 16.0 & 26.0 & 28.0 & 5.0 & 33.0 & \textbf{48.0} \\
Blocks Ranking Try & \(M(n)\) & 10.0 & 0.0 & 6.0 & 1.0 & 18.0 & 17.0 & 48.0 & \textbf{87.0} \\
Cover Blocks & \(M(n)\) & 0.0 & 0.0 & 0.0 & 2.0 & \textbf{68.0} & 0.0 & 42.0 & 56.0 \\
Press Button & \(M(n)\) & 0.0 & 0.0 & 0.0 & 0.0 & 0.0 & 0.0 & 19.0 & \textbf{34.0} \\
\(M(n)\) Average & -- & 5.0 & 4.8 & 5.5 & 7.3 & 28.5 & 5.5 & 35.5 & \textbf{56.3} \\
\midrule
Overall Average & -- & 5.8 & 5.9 & 10.4 & 9.8 & 42.0 & 4.0 & 34.8 & \textbf{69.8} \\
\bottomrule
\end{tabular}}
\end{table*}

\methodname{} achieves the highest reported success rate on eight of the nine RMBench tasks shown in Table~\ref{tab:rmbench_main} and exceeds benchmark-reported Mem-0 by \(27.8\) percentage points in overall average success. In the training-matched comparison with LingBot-VA, it improves the average success rate across all tasks from \(34.8\%\) to \(69.8\%\), with gains on both \(M(1)\) (\(34.2\%\) to \(80.6\%\)) and \(M(n)\) (\(35.5\%\) to \(56.3\%\)).

\subsection{Real-World Experiments}

\begin{figure}[t]
  \centering
  \includegraphics[width=0.98\linewidth]{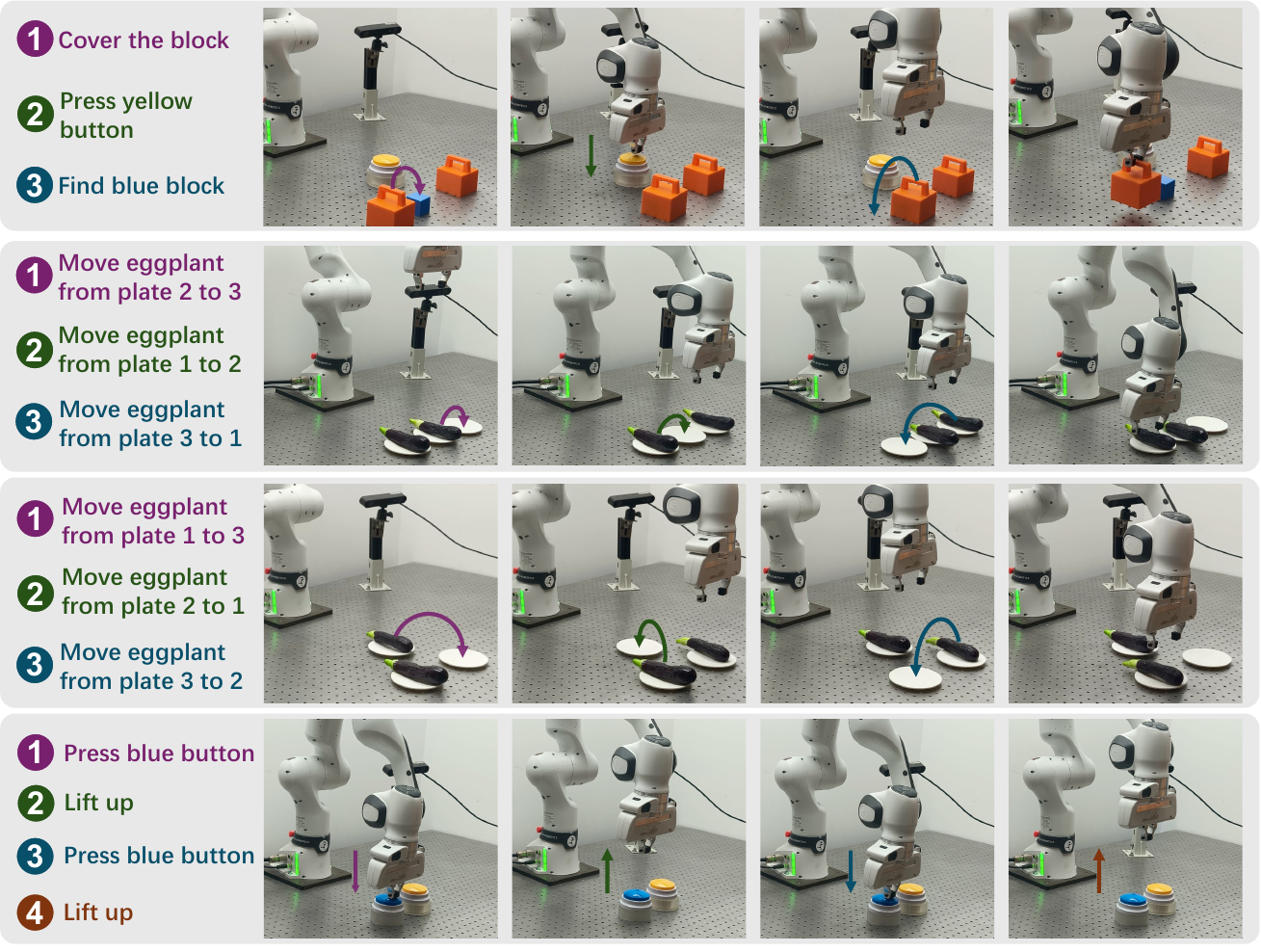}
  \caption{Real-world tasks, listed in the order used in Table~\ref{tab:real_results}}
  \label{fig:real_world_tasks}
\end{figure}

We use four multi-stage Franka Panda tasks requiring prior spatial state, event order, or target identity: Find Blue Block, Line Swap, Triangle Swap, and Press Twice (Fig.~\ref{fig:real_world_tasks}).

For each task, \(\pi_{0.5}\)~\cite{arxiv2504.16054}, Fast-WAM~\cite{arxiv2603.16666}, LingBot-VA~\cite{arxiv2601.21998}, and \methodname{} are trained using the same set of 15--25 demonstrations and evaluated in 10 full-task trials. Stage success ratio (SSR) is the number of completed stages divided by the total number of evaluated stages, and full-task success rate (SR) is the fraction of trials that complete the task. Table~\ref{tab:real_results} reports both metrics as counts, with averages in percent.

\begin{table*}[t]
\centering
\caption{Real-world Franka Panda results. Task entries are success counts over total attempts. SSR denominators are 40 for Find Blue Block, 60 for each swap task, and 30 for Press Twice; all SR denominators are 10. Average values are percentages.}
\label{tab:real_results}
\resizebox{0.75\textwidth}{!}{
\begin{tabular}{ccccccccccc}
\toprule
\multirow{2}{*}{Method} & \multicolumn{2}{c}{Find Blue Block} & \multicolumn{2}{c}{Line Swap} & \multicolumn{2}{c}{Triangle Swap} & \multicolumn{2}{c}{Press Twice} & \multicolumn{2}{c}{Avg. (\%)} \\
\cmidrule(lr){2-3}\cmidrule(lr){4-5}\cmidrule(lr){6-7}\cmidrule(lr){8-9}\cmidrule(lr){10-11}
 & SSR & SR & SSR & SR & SSR & SR & SSR & SR & SSR & SR \\
\midrule
\(\pi_{0.5}\) & \(0/40\) & \(0/10\) & \(3/60\) & \(0/10\) & \(0/60\) & \(0/10\) & \(0/30\) & \(0/10\) & 1.3 & 0.0 \\
Fast-WAM & \(0/40\) & \(0/10\) & \(0/60\) & \(0/10\) & \(0/60\) & \(0/10\) & \(0/30\) & \(0/10\) & 0.0 & 0.0 \\
LingBot-VA & \(21/40\) & \(1/10\) & \(47/60\) & \(6/10\) & \(31/60\) & \(4/10\) & \(\mathbf{30/30}\) & \(\mathbf{10/10}\) & 70.6 & 52.5 \\
\midrule
\methodname{}(ours) & \(\mathbf{39/40}\) & \(\mathbf{9/10}\) & \(\mathbf{55/60}\) & \(\mathbf{9/10}\) & \(\mathbf{51/60}\) & \(\mathbf{8/10}\) & \(\mathbf{30/30}\) & \(\mathbf{10/10}\) & \textbf{93.5} & \textbf{90.0} \\
\bottomrule
\end{tabular}}
\end{table*}

\methodname{} raises the average SSR from \(70.6\%\) to \(93.5\%\) and the average SR from \(52.5\%\) to \(90.0\%\). The largest SR gains occur on Find Blue Block (\(+80.0\) points; \(10.0\%\) to \(90.0\%\)) and Triangle Swap (\(+40.0\) points; \(40.0\%\) to \(80.0\%\)), which require cross-stage identity or spatial memory. The \(\pi_{0.5}\) and Fast-WAM baselines achieve no full-task successes.

Qualitative inspection reveals a shared failure mode of \(\pi_{0.5}\) and Fast-WAM: confusion about task progress when local observations are visually similar. In our demonstrations, initial and late-stage observations can appear nearly identical yet require different actions; selecting an action for the wrong stage can therefore cause task failure.

Press Twice is included for completeness but exhibits a ceiling effect: the local window covers approximately \(85\%\) of each demonstration, and both LingBot-VA and \methodname{} reach \(100\%\) full-task success. We therefore do not interpret this task as evidence of a memory-specific advantage. Given 10 trials per task, these real-world results are point estimates.

\subsection{Ablation Studies}

Table~\ref{tab:ablations} ablates bank structure and progress supervision on \texttt{Swap T} and \texttt{Swap Blocks}.

\begin{table}[t]
\centering
\caption{Ablation study on memory structure and progress supervision. Values are success rates in percent.}
\label{tab:ablations}
\small
\renewcommand{\arraystretch}{1.08}
\setlength{\tabcolsep}{2.8pt}
\begin{tabular}{@{}lccc@{}}
\toprule
Variant & Swap T & Swap Blocks & Avg. \\
\midrule
\(1\times32\) memory & 71.0 & 80.0 & 75.5 \\
\(4\times8\) memory & 88.0 & 92.0 & 90.0 \\
\(8\times12\) memory & \textbf{97.0} & \textbf{96.0} & \textbf{96.5} \\
\(8\times12\) w/o prog head & 92.0 & 89.0 & 90.5 \\
\bottomrule
\end{tabular}
\end{table}

At a fixed total capacity of 32 slots, replacing one bank with four banks improves the average success rate from \(75.5\%\) to \(90.0\%\), supporting the benefit of multi-bank structure. Increasing the configuration to \(8\times12\) further raises the average to \(96.5\%\), although this comparison changes both the number of banks and total capacity. Removing the progress head from the \(8\times12\) model reduces the average to \(90.5\%\).

\subsection{Cross-Bank Behavior Analysis}

\begin{figure}[t]
  \centering
  \subfloat[Retained memory events over task progress.\label{fig:cross_bank_timeline}]{
    \includegraphics[width=0.96\linewidth,height=0.22\textheight,keepaspectratio]{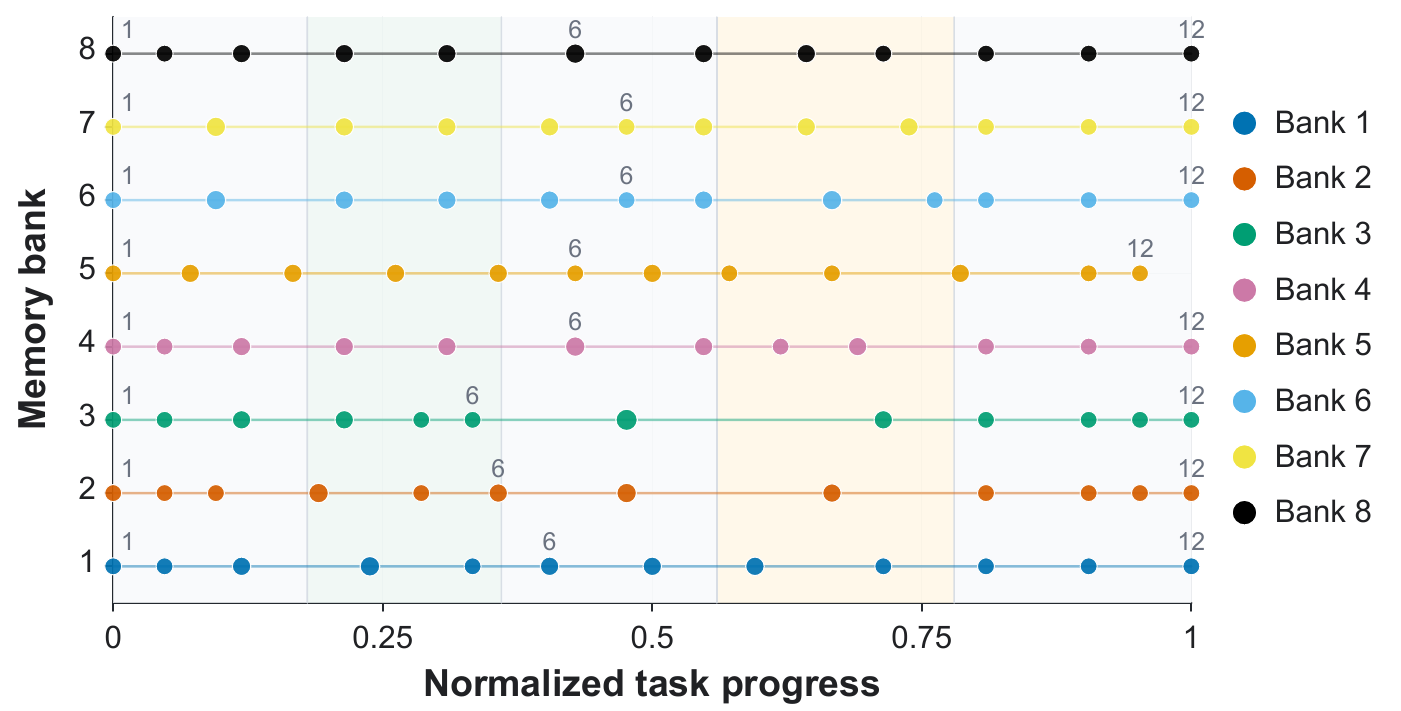}}
  \vspace{0.35em}
  \subfloat[Principal component analysis (PCA) of event tokens.\label{fig:cross_bank_pca}]{
    \includegraphics[width=0.96\linewidth,height=0.22\textheight,keepaspectratio]{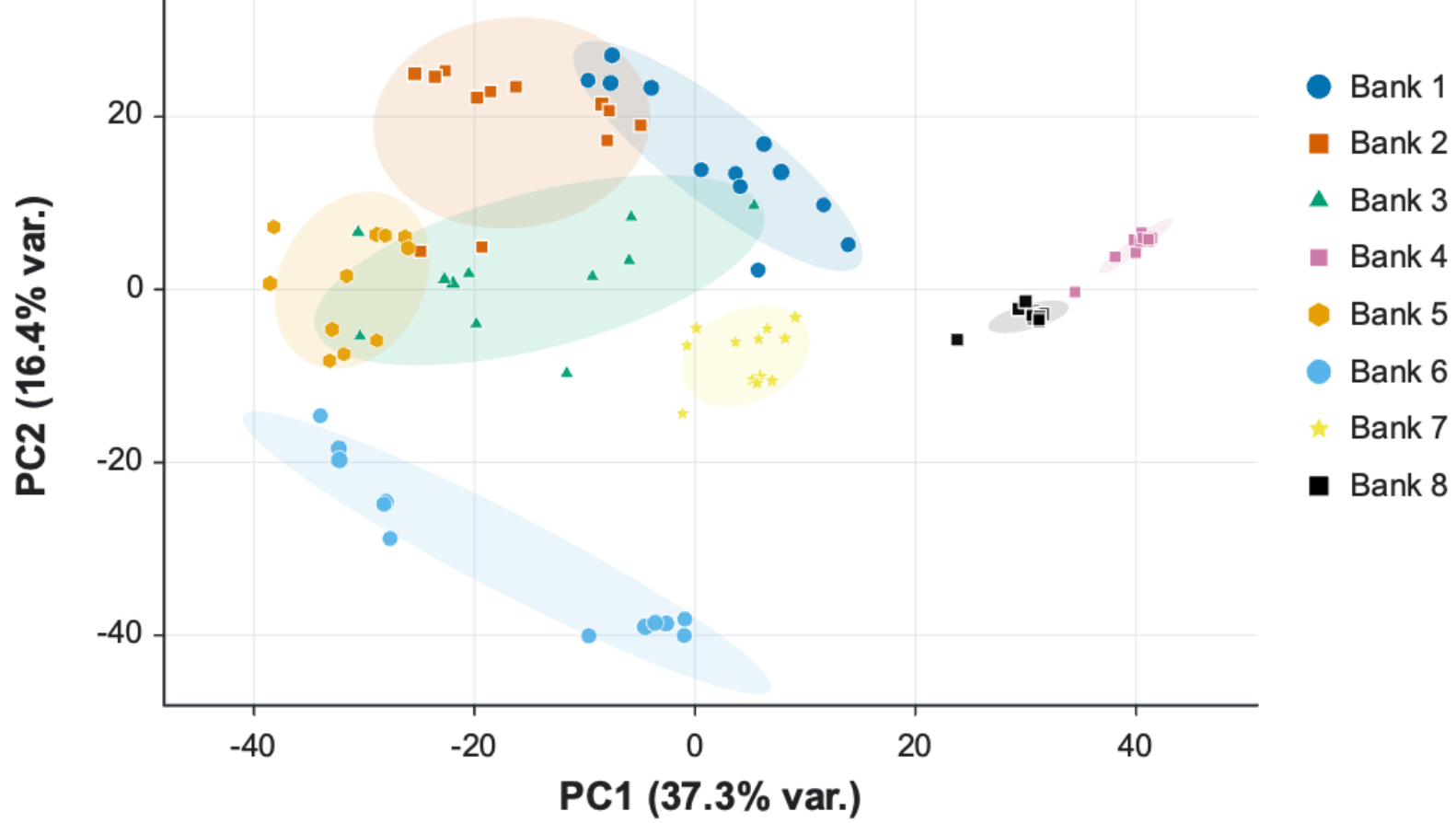}}
  \caption{Cross-bank behavior in one successful episode: (a) retained memory events over task progress and (b) PCA of event tokens by bank.}
  \label{fig:cross_bank_behavior}
\end{figure}

Fig.~\ref{fig:cross_bank_behavior} qualitatively illustrates the retained events and event-token distributions in one successful \texttt{put\_back\_block} episode.

The visualization is consistent with, but does not by itself establish, complementary cross-bank behavior. In this episode, banks retain different event positions, retained mass often concentrates near transitions, and PCA shows distinct but overlapping bank distributions; these observations do not establish a general trend.


\section{Conclusion}

\methodname{} is a memory-augmented WAM with DHEM for long-horizon robot manipulation. It reaches a \(90.0\%\) average full-task success rate across four real-world memory-dependent tasks.

\textit{Limitations and future work.} DHEM is evaluated primarily on memory-dependent manipulation tasks, so its effectiveness across broader task distributions, robot embodiments, and observation settings remains to be established. Although the ablations support multi-bank organization and progress supervision, they do not independently isolate every component of the memory-maintenance strategy. Progress supervision also relies on a coarse trajectory-position proxy rather than semantic task states, which may be less reliable when stage durations vary substantially. Moreover, the qualitative analysis provides only episode-level evidence of differentiated retention and does not establish stable semantic specialization across banks, tasks, or episodes. Future work will investigate adaptive memory mechanisms, richer progress representations, and broader real-world evaluations.

\bibliographystyle{IEEEtran}
\bibliography{references}

\end{document}